\documentclass[10pt,twocolumn,letterpaper]{article}

\usepackage{iccv}
\usepackage{times}
\usepackage{epsfig}
\usepackage{graphicx}
\usepackage{amsmath}
\usepackage{amssymb}
\usepackage{multirow}
\usepackage{caption} 
\usepackage{enumitem}

\DeclareMathOperator*{\argmax}{arg\,max}

\def\eg{\emph{e.g}\onedot} 
\def\ie{\emph{i.e}\onedot}

\def\etal{\emph{et al}\onedot}

\newcommand\blfootnote[1]{%
  \begingroup
  \renewcommand\thefootnote{}\footnote{#1}%
  \addtocounter{footnote}{-1}%
  \endgroup
}

\newlength\savewidth\newcommand\shline{\noalign{\global\savewidth\arrayrulewidth
  \global\arrayrulewidth 1pt}\hline\noalign{\global\arrayrulewidth\savewidth}}
  
\usepackage[pagebackref=true,breaklinks=true,letterpaper=true,colorlinks,bookmarks=false]{hyperref}

\iccvfinalcopy

\begin{document}

\title{ClawCraneNet: Leveraging Object-level Relation for \\ Text-based Video Segmentation}

\author{
  Chen Liang \quad Yu Wu \quad Yawei Luo \quad Yi Yang \\ 
  \small{Zhejiang University \quad  Baidu Research \quad ReLER, University of Technology Sydney}\\
}

\maketitle
\ificcvfinal\thispagestyle{empty}\fi

\begin{abstract}
\blfootnote{Extended version published in~\cite{liang2023locater}.} Text-based video segmentation is a challenging task that segments out the natural language referred objects in videos. It essentially requires semantic comprehension and fine-grained video understanding. Existing methods introduce language representation into segmentation models in a bottom-up manner, which merely conducts vision-language interaction within local receptive fields of ConvNets. 
We argue such interaction is not fulfilled since the model can barely construct region-level relationships given partial observations, which is contrary to the description logic of natural language/referring expressions.
In fact, people usually describe a target object using relations with other objects, which may not be easily understood without seeing the whole video.
To address the issue, we introduce a novel top-down approach by imitating how we human segment an object with the language guidance.
We first figure out all candidate objects in videos and then choose the refereed one by parsing relations among those high-level objects.
Three kinds of object-level relations are investigated for precise relationship understanding, \ie, positional relation, text-guided semantic relation, and temporal relation.
Extensive experiments on A2D Sentences and J-HMDB Sentences show our method outperforms state-of-the-art methods by a large margin.

\end{abstract}

\section{Introduction}
With the significant progresses achieved on computer vision and natural language processing, novel tasks requiring a joint understanding of both visual and linguistic modalities emerge recently, \eg, visual question answering \cite{anderson2018bottom,lu2016hierarchical,antol2015vqa}, image \cite{you2016image,anderson2018bottom,wu2018decoupled} and video captioning \cite{gao2017video,wang2018video}, vision-dialog navigation \cite{anderson2018vision,zhu2020vision} and so on. Inspired by such great success, Gavrilyuk~\etal~\cite{gavrilyuk2018actor} introduces a challenging task of text-based video segmentation, which takes a video and a natural language description as inputs, and prospects a set of segmentation masks for the referent.
Certain solutions for tackling this task lie in a comprehensively understanding of the visual and linguistic information with a fine-grained analysis of video contents.

\begin{figure}[t]
\begin{center}
     \includegraphics[width=1\linewidth]{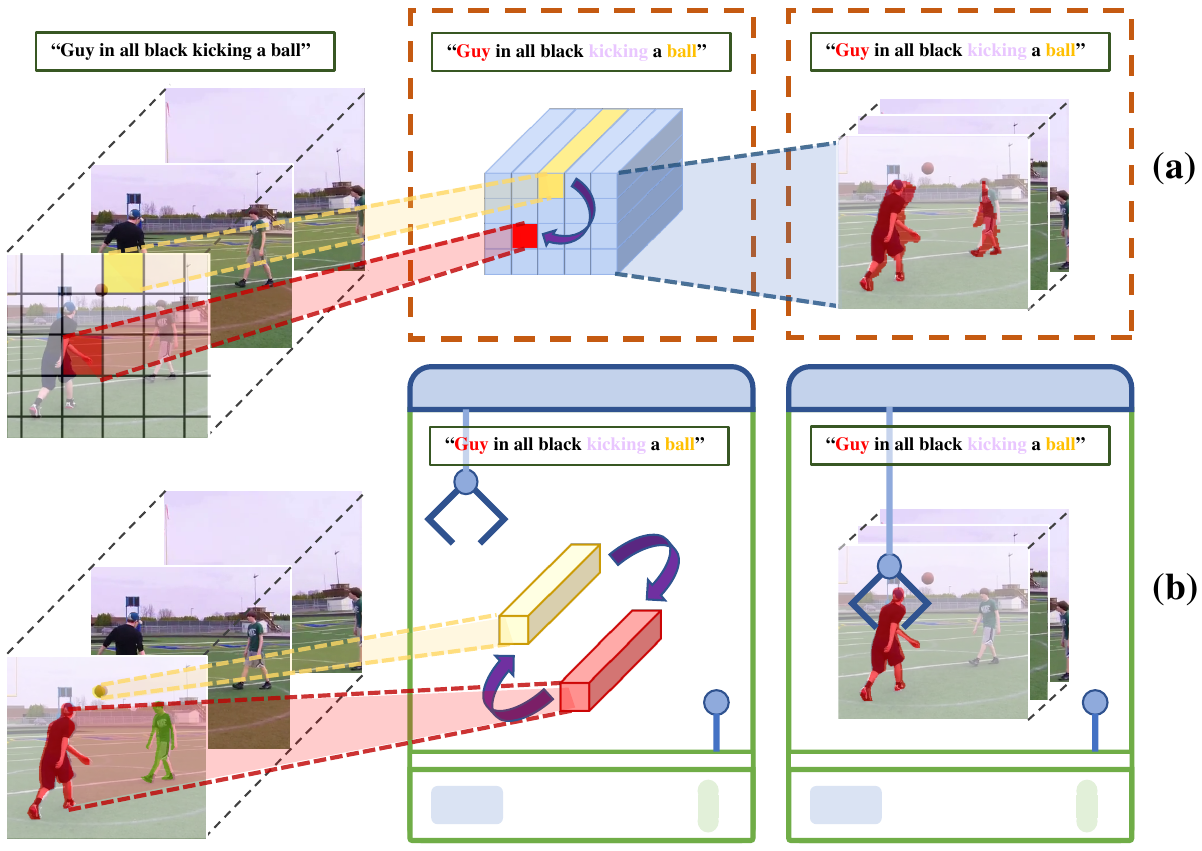}
\end{center}
\vspace{-12pt}
\captionsetup{font=small}
      \caption{\small 
      (a) Previous \textit{bottom-up methods} mainly perform semantic relationship formulation at the pixel level. Corresponding models could not correctly identify the high-level relation merely based on local perceptive fields, and directly leads to an ambiguous prediction.
      (b) Our \textit{top-down pipeline} first performs feature extraction for objects and then model the crucial relation information based on a high-level sensation, leading to better segmentation masks by conducting multi-modal retrieving.
      Vividly, we analogize the process of retrieving a visual object with the linguistic query as playing a \textbf{claw crane} machine. 
      }\label{fig:motivation}
\vspace{-8pt}
\end{figure}

Accordingly, several recent works \cite{wang2019asymmetric,mcintosh2020visual,wang2020context,ningpolar} proposed to capitalize on the pixel-level visual language relationship for video comprehension. These techniques are approached by a bottom-up paradigm, which mainly focuses on multi-modal feature fusion and pixel-level relation establishment to generate the segmentation masks directly. 
While some effectiveness has been achieved, these attempts generally lack the sensation of object-level information and relation, 
which is crucial for understanding semantic information, especially for multi-modal comprehension \cite{hu2020bi,huang2020referring}. 
As illustrated in Figure \ref{fig:motivation} (a), methods with low-level sensation could only formulate local region-level relationship, which is insufficient for modeling semantic relationship with natural language description logic.
Thus the bottom-up approaches would inevitably introduce noisy relationship modeling and inaccurate object comprehension, leading to ambiguous segmentation results.

To tackle the above issue, partially inspired by the human visual cognitive system that humans preferentially direct attention towards meaningful entities (object-orientated)~\cite{henderson2017meaning,buswell1935people,wolfe2017five}, and then extracts structured information on how entities relate to or interact with each other (relation-based) \cite{kalkstein2020person,wolfe2017five,kao2005neural}, 
we propose a top-down pipeline to mimic how humans localize the referent with language guidance, \ie, first finding candidates and then parsing relations.
Specifically, by applying an off-the-shelf instance segmentation module to find out candidate objects, we could tackle the Text-based Video Segmentation problem in an object-level cross-modal retrieval manner.
To our best knowledge, this is the first attempt to tackle the text-based video segmentation problem from the top-down view.

We further explore three kinds of object-level relations in our top-down pipeline, \ie, positional relation, semantic relation, and temporal relation. 

Firstly, we propose a \textit{relative position encoding} module to encode spatial information of each candidate object.  
On the basis of absolute position, we further consider the relative ranked index for each object according to its coordinates.
The relative index addresses the spatial relations that are common in natural language descriptions, \eg, ``the second guy from the left".

Secondly, we propose a language-guided object attention module to construct \textit{semantic relations}, which directly highlights the referring entities.
As shown in Figure~\ref{fig:motivation}, related objects (\eg, ``guy" and ``ball") will exchange information according to responsiveness to relational expressions (\eg, ``kicking").
After relation-aware language comprehension, a particular referent feature (\eg, ``guy") would contain rich semantic evidence including relationship (\eg, ``kicking a ball") and attribute (\eg, ``in black").

Thirdly, we investigate the \textit{temporal relationship} between inter-frame objects by a merge-by-track diagram.
Particularly, with a multi-object tracking strategy, we perform inter-frame object association based on similarities, and build temporally related tracks with the Hungarian algorithm \cite{kuhn1955hungarian}. 
Finally, the final prediction is performed according to the average of confidence scores in each track.

In this way, we obtain a visual embedding with rich individual and mutual information, 
which would facilitate a correct language-to-vision corresponding in the complex video context.  
Vividly, these visual objects are like well-packaged dolls displayed in a claw crane machine, and the linguistic description performs as a claw looking for the shiniest doll. Our network just explicitly formulates the retrieving pipeline, and that's why we named our network ClawCraneNet.
The main contributions are as follows:
\begin{itemize}[leftmargin=*]
\item We propose a novel top-down pipeline that tackles the text-based video segmentation task in a retrieval manner.

\item  
We explicitly investigate three kinds of object-level relations to progressively construct discriminative visual embedding, \ie, relative positional relation, cross-object semantic relation, and inter-frame temporal relation. 

\item The proposed method significantly outperforms state-of-the-art methods on two popular text-guided video segmentation datasets, \ie, A2D, and J-HMDB.
\end{itemize}

\section{Related Work}
\subsection{Referring Image Segmentation}
Referring expression segmentation aims at precisely localizing the entity referred by a natural language expression with a pixel-level segmentation mask. 
The bottom-up methods \cite{hu2016segmentation,liu2017recurrent,ye2019cross,qiu2019referring,chen2019referring,margffoy2018dynamic,huang2020referring,hui2020linguistic,chen2019see,yang2019fast} mainly construct a multi-modal feature, then generate referring masks after some refinement progress. Most state-of-the-art works conduct the structures of fully convolutional network (FCN) \cite{long2015fully} to generate the pixel-level segmentation mask. 
At first, Hu \etal \cite{hu2016segmentation} directly leverage the concatenation of visual and linguistic features from CNN and LSTM to construct multi-modal feature and generate the final mask. Later, several techniques are incorporated into this field 
, \eg multi-modal LSTM \cite{liu2017recurrent}, image-to-word attention \cite{ye2019cross}, dynamic filter \cite{margffoy2018dynamic}, and adversarial learning \cite{qiu2019referring} or cycle-consistency \cite{chen2019referring} between referring expression and its reconstructed caption.
Recently, to explore the relationship between multi-modal features~\cite{wu2019DAM} and further model the structural context, Hu \etal \cite{hu2016segmentation} propose a bi-directional cross-modal attention module to emphasize visual guidance on linguistic features. Huang \etal \cite{huang2020referring} utilize a graph-based structure to progressively exploits different types of words in the expression. 
Different from these works which focus on low-level feature comprehension. Inspired by the human vision system, \eg finding the referring objects then parsing the relation, we investigate object-level feature retrieving as another alternative.

\subsection{Top-down Text-based Object Grounding}
The existing top-down methods \cite{zhang2020does, sadhu2020video, yang2020grounding, yu2018mattnet,wu2020phrasecut,bajaj2019g3raphground} mainly leverage the pre-trained detector, \eg, Mask R-CNN \cite{he2017mask}, to generate object proposals, and then rank the box-level objects according to similarity score among vison-language embeddings. 
With the same attempts, we tend to follow the same top-down strategy, utilize an off-the-shelf instance segmentation method to perceive candidate objects.
However, different from existing methods which mainly realize the box-level object feature matching as the main task and consider the segmentation mask as a by-product of the modular comprehension procedure by simply replacing the output heads.
With the object feature constructed on bounding boxes, these methods would not handle the occlusions among objects in a video, especially for crowded scenes.
In this work, we try to directly model the object feature based on fine-grained segmentation masks to learn more discriminative object features.
Additionally, we further explore multi-modal relationship modeling among high-level visual object features.

\subsection{Text-based Video Segmentation}
Certain success has been achieved in referring image segmentation. Beyond image domain, the temporal coherence of referred video objects is still waiting to be explored. 
Recently, Gavrilyuk \etal \cite{gavrilyuk2018actor} extend Actor-Action Dataset (A2D) with human-annotated sentences and introduce the challenging task of actor and action video segmentation from referring expressions. They adopt language guided dynamic convolution filters to fuse the multi-modal feature. Since then, bottom-up methods have sprung up.
Wang \etal \cite{wang2019asymmetric} utilize asymmetric attention mechanisms to facilitate visual guided linguistic feature learning. Later, they \cite{wang2020context} extend vanilla dynamic convolution with a context modulated dynamic convolution kernel. Ning \etal \cite{ningpolar} convert spatial relations to terms of direction and range for better linguistic spatial formulation.
McIntosh \etal \cite{mcintosh2020visual} introduce a capsule-based approach for better capturing the relationship between multi-modal features. For further mining continuous temporal information, they extend the A2D dataset with annotations for all frames. Hui \etal~\cite{hui2021collaborative} introduce an additional 2D spatial encoder to alleviate the intrinsic spatial misaligned problem in 3D CNNs. 

In this work, with the concrete objects obtained with the instance segmentation module, we explicitly exploit temporal coherence among all frames in a video including annotated key-frames and unlabeled frames.

\begin{figure*}[t]
\begin{center}
    \includegraphics[width=1\linewidth]{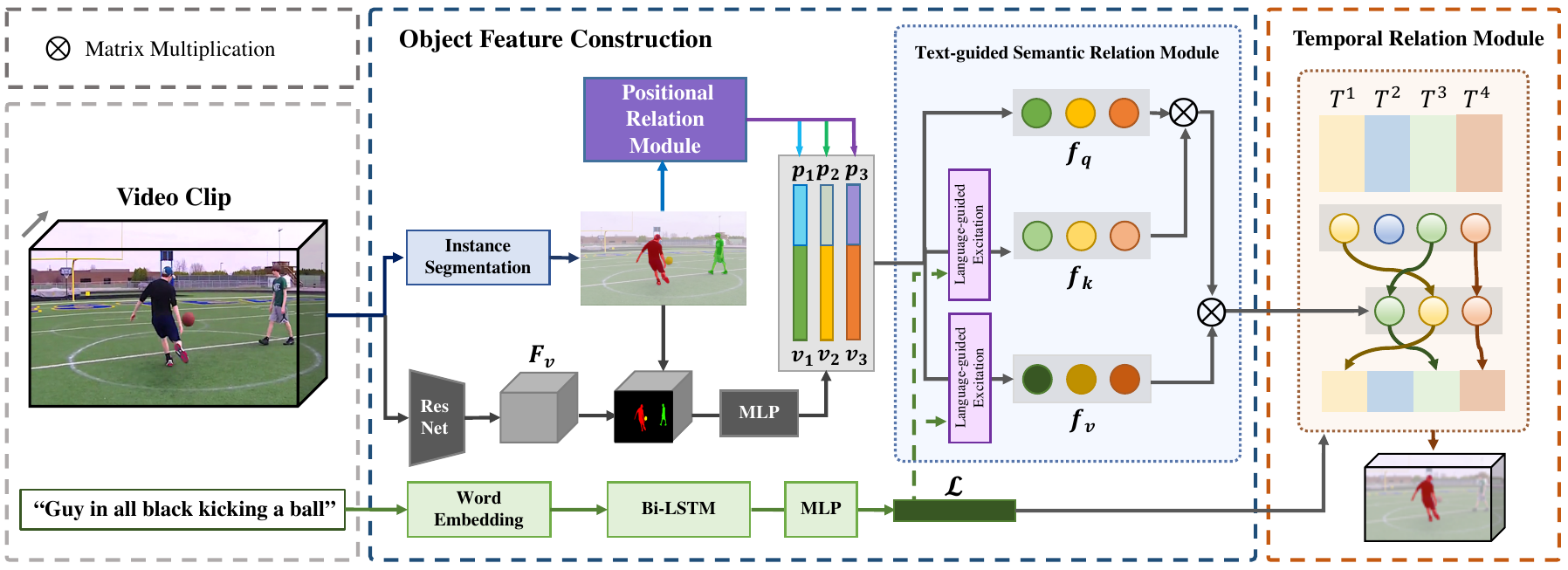}
\end{center}
\vspace{-16pt}
\captionsetup{font=small}
     \caption{\small\textbf{The framework of our proposed ClawCraneNet.} 
     As a top-down pipeline, objects are first perceived by an off-the-shelf instance segmentation module, and then selected by finding best visual-semantic match. During this process, we populate information among objects by performing three kinds of relation formulation module, \ie, positional relation, text-guided semantic relation, and temporal relation. We then utilize linguistic embedding to retrieve the final prediction.
     }\label{fig:network_main}
\vspace{-8pt}
\end{figure*}

In this work, with the concrete objects obtained with the instance segmentation module, we explicitly exploit temporal coherence among all frames in a video including annotated key-frames and unlabeled frames.

\section{Methods}

\subsection{Top-Down Pipeline}
Previous work \cite{wang2019asymmetric,mcintosh2020visual,wang2020context,ningpolar} tackles Text-based Video Segmentation task from a bottom-up pixel-level way. 
Differently, we view it as a cross-modal retrieval problem by decomposing the task into two stages.
The first stage is to find out all potential candidate objects and their masks in the video, and the second one is to populate information among objects and select the best matched candidate given the referring sentence. 

Suppose a video $V$ has $M$ frames, where each frame $f_i$ contains $k_i$ candidate objects $\{x_i^1,x_i^2,...,x_i^{k_i}\}$.
Our target is to retrieve a set of referring objects $T^* = \{x_1^*, x_2^*,...,x_M^*\}$ in a video by a natural language referring sentence $L$. The sentence $L$ is composed of a sequence of words $(w_1,w_2,...,w_{N_l} )$, where $N_l$ is the length of input sentence.
The task is taking the language description $L$ to retrieve the target object track $T$ from a video.

\vspace{1mm}
\noindent
\textbf{Linguistic Embedding Construction.}
The language expression $L$ is first processed via a bi-LSTM \cite{huang2015bidirectional}, where the hidden states $\{h_1, h_2, \cdots h_{N_l}\}$ are further encoded by a self-guided attention module. In particular, the linguistic embedding $\mathcal{L}$ can be obtained by:
\begin{align}
\mathcal{L} &= \texttt{MLP}(\sum_{i=1}^{N_l}{\alpha_{i}h_i}),
\end{align}
where $\alpha$ is the word-level attention weights that calculated by $\alpha_i = \texttt{softmax}(\texttt{fc}(h_i))$ and $\texttt{MLP}$ denotes the multi-layer perception.
The self-guided attention module introduces a flexible way for the language encoder to focus on keywords and reduce the negative impact caused by sentence truncation or padding.
Following we illustrate our designed top-down pipeline for this task.

\vspace{1mm}
\noindent
\textbf{Mask Out Foreground Objects.}
To localize objects in videos, we first build an instance segmentation model by considering the visual content only.
Specifically, we use CondInst~\cite{tian2020conditional} as our backbone, and train the model using all the object masks.
Then we apply the instance segmentation model on each frame and detect and segment all the foreground objects as candidates.
Note that we do not exploit additional data/annotations via the instance segmentation model.
Denote the object segmentation masks in frame $f_i$ as $\{o^j\}_{j=1}^{N_v}$, where $N_v$ is the number of candidates.

\vspace{1mm}
\noindent
\textbf{Individual Object features.}
We then obtain the $j$-th individual object feature $v^j$ by max-pooling on mask-cropped feature map extracted from the visual CNN model. 
Formally, the process could be achieved by,
\begin{align} \label{eq:individual_feat}
v^j &= \texttt{MLP}(\texttt{Max}(F_v \odot o^j)),
\end{align}
where $\odot$ is element-wise multiplication, $\texttt{Max}$ stands for global max pooling and $F_v$ denotes feature map of the entire frame. During training, the CNN model is updated in an end-to-end manner.
Via Eq.~\ref{eq:individual_feat}, we build individual object feature $v_j$ by applying its instance segmentation mask $o_j$ on top of the CNN feature map.

\vspace{1mm}
\noindent
\textbf{Find Visual-Linguistic Match.}
From the top-down perceptive, the final target is to select a best match among all candidate objects given the language input. Thus we train our model to maximize the matching score between referring object track and the language representation by,
\begin{align}
T^* &= \argmax \sum_{i=1}^M {S(v_i^*, \mathcal{L})}.
\end{align}
Therefore, the core of the problem is to learn a proper visual object embedding that distinguishes the target objects from others.
In the later sections, details about how to learn discriminative multi-modal embeddings are introduced.

\subsection{Object-level Visual Embedding Construction}
As many entities exist in a visual scene, semantic information of describing entity categories is not enough for distinguishing them. Therefore, it is natural to populate information among all the candidates to facilitate retrieving progress. 
In this section, on the basis of individual object representation, we further leverage three kinds of object-level relation, \ie,  positional relation, text-guided semantic relation, and inter-frame temporal relation.

\vspace{1mm}
\noindent
\textbf{Positional Relation Module.}
Positional information is crucial in depicting a object in images/videos.
We design a Positional Relation Module (PRM) to encode the object-level spatial information with $p_i = ( x_{min}^i, y_{min}^i, x_{max}^i, y_{max}^i, x_c^i, y_c^i, w_i, h_i, r_i^x, r_i^y )$
, where $(x_{min}^i, y_{min}^i)$, $(x_{max}^i, y_{max}^i)$, $(x_c^i, y_c^i)$, $w_i$ and $h_i$ are the normalized top-left coordinates, bottom-right coordinates, center coordinates, width and height of the smallest circumscribed box of segment $o_i$, respectively. 
The last two dimensions $r_i^x$ and $r_i^y$ are the normalized relative position index according to the x-axis and y-axis coordinates. 
Then, spatial enhanced object features $\mathcal{V}_i$ is calculated from: 
\begin{align}
\mathcal{V}_i = v_i + W_p(p_i), 
\end{align}
where $W_p$ is a learnable matrix.
With the explicitly modeling of relative position information, our network earns the ability to handle the referring like ``the second from the left" which is hard for low-level networks to infer with pixel-level spatial encoding only. 
We show by experiments that even with such slight guidance, our model does learn to comprehend relative spatial descriptions.
During the training phase, to enhance the model to be position-aware, we randomly horizontal flip the frame image and swap the corresponding direction textual descriptions (\eg, changing from ``right" to ``left").

\begin{figure}[t!]
\begin{center}
     \includegraphics[width=0.8\linewidth]{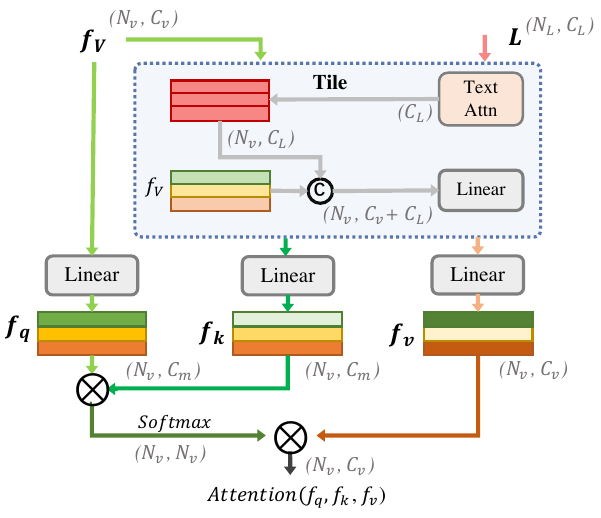}
\end{center}
\vspace{-16pt}
\captionsetup{font=small}
     \caption{\small\textbf{Illustration of the text-guided semantic relation module.} $\otimes$: Matrix Multiplication; \textcircled{c}: Matrix Concatenation; 
      Three boxes with different colors stand for three different objects. 
      Self-guided linguistic context $L$ is used as a guidance to infer the relationship between object features $f_V$. 
      }\label{fig:method_context}
\vspace{-8pt}
\end{figure}

\vspace{1mm}
\noindent
\textbf{Text-guided Semantic Relation Module.}
On formulating intra-frame object-level relationships, a simple idea is to employ the common-used vanilla attention module as a relation formulator.
However, vanilla attention boosts information exchange naively based on feature similarity, which is hard to distinguish with the within-modal similarity.
To diminish the gap, we introduce text guidance into the semantic relation module. As the referring example illustrated in Figure \ref{fig:motivation}, comparing with directly forming the relationship between ``Guy" and ``Ball", it is easier for the network to infer the ``Ball'' if it has already known that the ``Guy" is ``Kicking" something.
Based on the aforementioned motivation, we devise a Text-guided Semantic Relation Module (TSRM) to leverage relational expression in language description.
As illustrated in Figure \ref{fig:method_context}, TSRM takes concatenated object features 
$f_V = (\mathcal{V}_1,\mathcal{V}_2,...,\mathcal{V}_{N_v})$ and the sentence $L$ as inputs.
TSRM first learns self-guided weights for representing relationship-aware linguistic feature $f_t$ by,
\begin{align}
f_o &= W_o(\texttt{Concat}(f_V,f_t)),
\end{align}
where $\texttt{Concat}(·,·)$ represents the concatenation operation along the channel axis, $W_o$ is a learnable matrix.
Next in relation stage, the original visual feature is utilized to query the multi-modal feature. 
Given the query $f_q = W_q f_V$, the key $f_k = W_k f_o$, and the value $f_v = W_v f_o$, the process to get text-guided object features $F_o=(\mathcal{V}_1^r,\mathcal{V}_2^r,...,\mathcal{V}_{N_v}^r)$ could be formulated as,
\begin{equation}
F_o = f_V + \texttt{softmax}(\frac{f_q f_k^T}{\sqrt{d_k}})f_v,
\end{equation}
where $\mathcal{V}_{i}^r$ is the relation enhanced object feature and $d_k$ is the channel dimension of $f_k$.
During the procedure, each object could earn query-focused global context information especially when there is a strong response between the visual object and linguistic description.

\vspace{1mm}
\noindent
\textbf{Temporal Relation Module.}
To deal with blurry or complicated scenes in a video, a natural idea is to borrow the confident judgment from a clear scene for tackling hard scenes.
In this part, we employ a tracking-based strategy to meet this purpose. 
Particularly, cross-frame objects are associated based on visual similarities to form a track.
Given any two object ($x_i$, $x_j$) from adjacent frames, the association similarity $\mathcal{S}_s$ is formulated as follows:
\begin{align}
\mathcal{S}_s(x_i, x_j) &= \mathcal{S}_c(\mathcal{V}_i^r,\mathcal{V}_j^r) + \alpha * U(x_i, x_j), \label{eq:asso}
\end{align}
where $\mathcal{S}_c$ denotes the cosine similarity and $U$ represents the mask IoU. 
Following the multi-object tracking strategy in~\cite{xu2020segment}, we maintain several active tracks initialized from the first frame by treating each candidate object as an exclusive track. For each frame, we compute the visual similarity between the all active tracks and all candidate object embeddings in the current frame according to Eq.~\ref{eq:asso}. The association procedure is allowed, only when the visual similarity is greater than a threshold $\gamma$ and the active track would be ended if it is not updated for $\beta$ matching rounds.
The Hungarian algorithm \cite{kuhn1955hungarian} is applied to perform multi-object matching. After the aforementioned procedure, unassigned segments will start new tracks and repeat the object association until the final frame. 
Since the target object may appear or disappear in internal frames, it's reasonable to allow intermittent tracks.

\begin{table*}
\small
\begin{center}
\resizebox{.95\textwidth}{!}{
     \setlength\tabcolsep{8pt}
     \renewcommand\arraystretch{1.0}
\begin{tabular}{l|ccccc|c|cc|c}
\hline
\multirow{2}{*}{\textbf{Methods}}       & \multicolumn{5}{c|}{\textbf{Overlap}}                                                 & \textbf{mAP} & \multicolumn{2}{|c|}{\textbf{IoU}} & \multirow{2}{*}{\textbf{FPS}}  \\
 & P@0.5 & P@0.6 & P@0.7 & P@0.8 & P@0.9 & 0.5:0.95 & Overall & Mean & \\ 
\shline
Hu \etal \cite{hu2016segmentation} \tiny ECCV16 & 34.8 & 23.6 & 13.3 & 3.3  & 0.1 & 13.2 & 47.4 & 35.0 & - \\
Li \etal \cite{li2017tracking} \tiny CVPR17 & 38.7 & 29.0 & 17.5 & 6.6  & 0.1 & 16.3 & 51.5 & 35.4 & - \\
Gavrilyuk \etal \cite{gavrilyuk2018actor} \tiny CVPR18 & 53.8 & 43.7 & 31.8 & 17.1 & 2.1 & 26.9 & 57.4 & 48.1 & - \\
Wang \etal \cite{wang2019asymmetric} \tiny ICCV19 & 55.7 & 45.9 & 31.9 & 16.0 & 2.0 & 27.4 & 60.1 & 49.0 & 8.64 \\
McIntosh \etal \cite{mcintosh2020visual} \tiny CVPR20 & 52.6 & 45.0 & 34.5 & 20.7 & 3.6 & 30.3 & 56.8 & 46.0 & - \\
Wang \etal \cite{wang2020context} \tiny AAAI20 & 60.7 & 52.5 & 40.5 & 23.5 & 4.5 & 33.3 & 62.3 & 53.1 & 7.18 \\ 
Ning \etal \cite{ningpolar} \tiny IJCAI20 & 63.4 & 57.9 & 48.3 & 32.2 & 8.3 & 38.8 & \textbf{66.1} & 52.9 & 5.42 \\ 
\hline
\textbf{Ours} & \textbf{70.4} & \textbf{67.7} & \textbf{61.7} & \textbf{48.9} & \textbf{17.1} & \textbf{49.4} & 63.1 & \textbf{59.9} & \textbf{9.27} \\
\hline
\end{tabular}
}
\end{center}
\vspace{-16pt}
\captionsetup{font=small}
\caption{\small Comparison with state-of-the-art methods on the A2D Sentences using \textit{IoU} and \textit{Precision@K} as metrics.
}
\vspace{-8pt}
\label{table:a2dsota}
\end{table*}

\begin{table*}
\small
\begin{center}
\resizebox{.95\textwidth}{!}{
     \setlength\tabcolsep{10pt}
     \renewcommand\arraystretch{1.0}
\begin{tabular}{l|ccccc|c|cc}
\hline
\multirow{2}{*}{\textbf{Methods}}        & \multicolumn{5}{c|}{\textbf{Overlap}}                                                                          & \textbf{mAP} & \multicolumn{2}{|c}{\textbf{IoU}} \\
 & P@0.5 & P@0.6 & P@0.7 & P@0.8 & P@0.9 & 0.5:0.95 & Overall & Mean \\ 
\shline
Hu \etal \cite{hu2016segmentation} \tiny ECCV16          & 63.3 & 35.0 & 8.5 & 0.2 & 0.0 & 17.8 & 54.6 & 52.8 \\
Li \etal \cite{li2017tracking} \tiny CVPR17             & 57.8 & 33.5 & 10.3 & 0.6 & 0.0 & 17.3 & 52.9 & 49.1 \\
Gavrilyuk \etal \cite{gavrilyuk2018actor} \tiny CVPR18   & 71.2 & 51.8 & 26.4 & 3.0 & 0.0 & 26.7 & 55.5 & 57.0 \\
Wang \etal \cite{wang2019asymmetric} \tiny ICCV19       & 75.6 & 56.4 & 28.7 & 3.4 & 0.0 & 28.9 & 57.6 & 58.4 \\
McIntosh \etal \cite{mcintosh2020visual} \tiny CVPR20   & 67.7 & 51.3 & 28.3 & 5.1 & 0.0 & 26.1 & 53.5 & 55.0 \\
Wang \etal \cite{wang2020context} \tiny AAAI20          & 74.2 & 58.7 & 31.6 & 4.7 & 0.0 & 30.1 & 55.4 & 57.6 \\ 
Ning \etal \cite{ningpolar} \tiny IJCAI20                & 69.1 & 57.2 & 31.9 & 6.0 & 0.1 & 29.4 & - & - \\ 
\hline
\textbf{Ours} & \textbf{88.0} & \textbf{79.6} & \textbf{56.6} & \textbf{14.7} & \textbf{0.2} & \textbf{43.3} & \textbf{64.4} & \textbf{65.5} \\
\hline
\end{tabular}
}
\end{center}
\vspace{-16pt}
\captionsetup{font=small}
\caption{\small Comparison with state-of-the-arts on the J-HMDB Sentences with the best model trained on A2D Sentences \textbf{without finetuning}.
}
\vspace{-8pt}
\label{table:jhmdbsota}
\end{table*}

\subsection{Training and Inference}
Once all the comprehension procedure has been done, the final set of visual embeddings $\{\mathcal{V}_{1}^r, \mathcal{V}_{2}^r, \cdots, \mathcal{V}_{N_v}^r \}$ and linguistic embedding $\mathcal{L}$ are obtained.
We calculate the cosine similarity between the linguistic embedding and each visual candidate, ${\mathcal{V}_{i}^r}^{\texttt{T}} \mathcal{L}$. Here we enforce all vectors to be L2-normalized feature embeddings, \ie, $||\mathcal{V}_{i}^r||=1$, $||\mathcal{L}||=1$.
We adopt the contrastive learning loss for optimizing the model,
\begin{align}
s_i &= \frac{\exp({\mathcal{V}_{i}^r}^{\texttt{T}} \mathcal{L}/\tau)}{\sum_{j=1}^{N_v} \exp({\mathcal{V}_{j}^r}^{\texttt{T}} \mathcal{L}/\tau)}, \label{eq:r2}\\
\texttt{loss} &= -\log(s_{gt}) \label{eq:r3},
\end{align}
where $s_{gt}$ is the matching score of the ground-truth object, $\tau$ is a temperature parameter that controls the concentration level of the distribution. Higher $\tau$ leads to a softer probability distribution. We set $\tau=0.1$ in our experiments.

During the inference phase, our network first extracts multi-modal embeddings for each frame. Then, Temporal Relation Module is conducted to obtain the candidate tracks. 
The final track is retrieved by choosing the candidate track with the highest matched candidate.

Besides, as a prerequisite for the object association step, visual embeddings belonging to the same object are implicitly pulled together since they are all expected to be close with the same linguistic embedding. 
More explanations are conducted in supplementary materials.

\setlength{\tabcolsep}{5pt}
\begin{table*}
\small
\begin{center}
\begin{tabular}{l|ccccc|c|cc}
\hline
\multirow{2}{*}{\textbf{Methods}}        & \multicolumn{5}{c|}{\textbf{Overlap}}                                                                          & \textbf{mAP} & \multicolumn{2}{|c}{\textbf{IoU}} \\
 & P@0.5 & P@0.6 & P@0.7 & P@0.8 & P@0.9 & 0.5:0.95 & Overall & Mean \\ 
\shline
Top-Down Pipeline & 64.4 & 62.1 & 56.9 & 45.3 & 16.4 & 45.6 & 58.4 & 55.1 \\
\hline
+ Absolute PE        & 66.8 & 64.2 & 58.7 & 46.9 & 16.4 & 47.0 & 60.6 & 56.9 \\
+ Relative PE        & 66.3 & 63.9 & 60.2 & 46.9 & 16.3 & 46.9 & 60.2 & 56.7 \\
+ PRM        & 67.2 & 64.5 & 58.9 & 47.2 & 16.4 & 47.2 & 61.1 & 57.4 \\
\hline
+ PRM + Vanilla-attention   & 67.5 & 64.6 & 59.2 & 47.3 & 16.6 & 47.3 & 61.0 & 57.6 \\
+ PRM + TSRM        & 68.6 & 66.0 & 60.2 & 48.1 & 16.8 & 48.3 & 62.3 & 58.6 \\ \hline
+ PRM + TSRM + TRM                            & 70.4 & 67.7 & 61.7 & 48.9 & 17.1 & 49.4 & 63.1 & 59.9 \\
\hline
\end{tabular}
\end{center}
\vspace{-16pt}
\captionsetup{font=small}
\caption{\small Ablation studies on A2D Sentences. PE indicates Position Encoding. Positional Relation Module, Text-guided Semantic Relation Module, and Temporal Relation Module are abbreviated as ``PRM", ``TSRM", and ``TRM'', respectively.
}
\vspace{-8pt}
\label{table:ablation}
\end{table*}

\section{Experiments}
\subsection{Datasets and Evaluation Criteria}
We conduct our experiments on two extended datasets: \textbf{A2D Sentences} and \textbf{J-HMDB Sentences}. These datasets are released in \cite{gavrilyuk2018actor} by additionally providing corresponding human natural descriptions on original A2D \cite{xu2015can} and J-HMDB \cite{jhuang2013towards} respectively.
\noindent\textbf{A2D Sentences} contains 3782 videos in total with 8 action classes performed by 7 actor classes. Each video in A2D has 3 to 5 frames annotated with pixel-level actor-action segmentation masks. Besides, it contains 6,655 sentences corresponding to actors and their actions. Following settings in \cite{wang2019asymmetric}, we split the whole dataset into 3017 training videos, 737 testing videos, and 28 unlabeled videos.
\noindent\textbf{J-HMDB Sentences} contains 928 short videos with 928 corresponding sentences describing 21 different action classes. Pixel-wise 2D articulated human puppet masks are provided for evaluating segmentation performance.

The proposed method is evaluated with the criteria of Intersection-over-Union (IoU) and precision. The overall IoU computes the ratio of the total intersection area divided by the total union area over testing samples. The mean IoU is the averaged IoU over all samples, which treats samples of different sizes equally. We also measure precision@K which considers the percentage of testing samples whose IoU scores are higher than threshold K at 5 different IoU thresholds and calculate mean average precision over 0.50:0.05:0.95 \cite{gavrilyuk2018actor}.

\subsection{Implementation Details}
Our network is built on a one-stage instance segmentation method named CondInst \cite{tian2020conditional} for balanced performance and speed. It could be replaced with any other instance segmentation network. This model is initialized from ResNet101 \cite{he2016deep} pre-trained on ImageNet \cite{deng2009imagenet} and further trained exclusively on A2D \cite{xu2015can}. Note that we do not leverage any additional data/annotations when building the instance segmentation module.

For visual and linguistic feature extractor, we adopt ResNet50 \cite{he2016deep} model pre-trained on ImageNet \cite{deng2009imagenet} as visual backbone and bi-LSTM \cite{huang2015bidirectional} as text encoder. All input frames are resized to $320 \times 320$.
Following the settings in \cite{gavrilyuk2018actor}, the maximum length of sentences is set to 20 and the dimension of word vector is 1000. We employ the hidden states of bi-LSTM \cite{huang2015bidirectional} as sentence features with a dimension of 2000. The word embeddings are initialized with one-hot vectors without any pre-trained weights applied. The cross frame entity association threshold $\gamma$ is set to $0.8$ by default. Training is done with Adam optimizer \cite{kingma2014adam} with an initial learning rate of $0.0001$, and a scheduler that waits for $2$ epochs after loss stagnation to reduce the learning rate by a factor of $10$. The batch size is $16$.

\subsection{Comparison with State-of-the-Art Methods}
We compare our ClawCraneNet with other state-of-the-art text-based video segmentation models following the settings in \cite{gavrilyuk2018actor} on the two datasets, \ie, A2D Sentences and J-HMDB Sentences. The comparison results are demonstrated in Table~\ref{table:a2dsota} and Table~\ref{table:jhmdbsota}.
First on A2D Sentences, we evaluate  \cite{hu2016segmentation,li2017tracking} pre-trained on ReferIt dataset \cite{kazemzadeh2014referitgame} and then fine-tuned version on A2D sentences. Other methods including ours are trained on A2D Sentences exclusively.
As shown in Table \ref{table:a2dsota}, with the help of object-level relation comprehension, our approach achieves state-of-the-art performance on most metrics with a remarkable margin, especially at higher IoU thresholds.
On $P@0.8$, our method outperforms the SOTA by a large margin of 16.7$\%$. 
Moreover, we bring 6.8$\%$ improvement on Mean IoU and 10.4$\%$ in mAP over SOTA respectively, which directly proves the effectiveness of our method.
In spite of such obvious achievement on mean IoU, we get relatively poor performance on overall IoU. Owing to the special favor of large objects, overall IoU lacks the perception for smaller objects which is crucial for reflecting model performance. It seems ClawCraneNet not only captures obvious larger objects but also learns the object-level semantic context for distinguishing small objects. 
Besides, we found our method is also efficient (high FPS) compared to other bottom-up methods.

On J-HMDB Sentences, for fair comparisons, we follow the setting in \cite{wang2019asymmetric,wang2020context,mcintosh2020visual,gavrilyuk2018actor}, and evaluate our model pre-trained on A2D sentences without any additional fine-tuning, which is kept the same as other compared methods.
Our approach significantly outperforms previous state-of-the-art methods on all metrics considered. 
For the result of $P@0.9$, one possible reason is that the ground truth masks from J-HMDB Sentences are generated from puppets. It's hard to fit the data distribution with a model trained from precise segmentation masks.

\begin{table}
\small
\begin{center}
\begin{tabular}{c|c|c|cc}
\hline
\multirow{2}{*}{\textbf{Methods}} & \multirow{2}{*}{\textbf{Backbone}} & \textbf{mAP} & \multicolumn{2}{|c}{\textbf{IoU}} \\
 & & 0.5:0.95 & Overall & Mean \\
\shline
CondInst & R-101-FPN & 49.4 & 63.1 & 59.9 \\
CondInst & R-50-FPN & 48.3 & 62.7 & 59.4 \\
Mask R-CNN & R-50-FPN & 48.1 & 62.4 & 58.8 \\
\hline
\end{tabular}
\end{center}
\vspace{-16pt}
\captionsetup{font=small}
\caption{\small Impact of Instance Segmentation Modules. With weaker modules, we still get better performance than SOTAs.}
\vspace{-8pt}
\label{table:ins}
\end{table}

\begin{figure*}[t]
\begin{center}
    \includegraphics[width=1\linewidth]{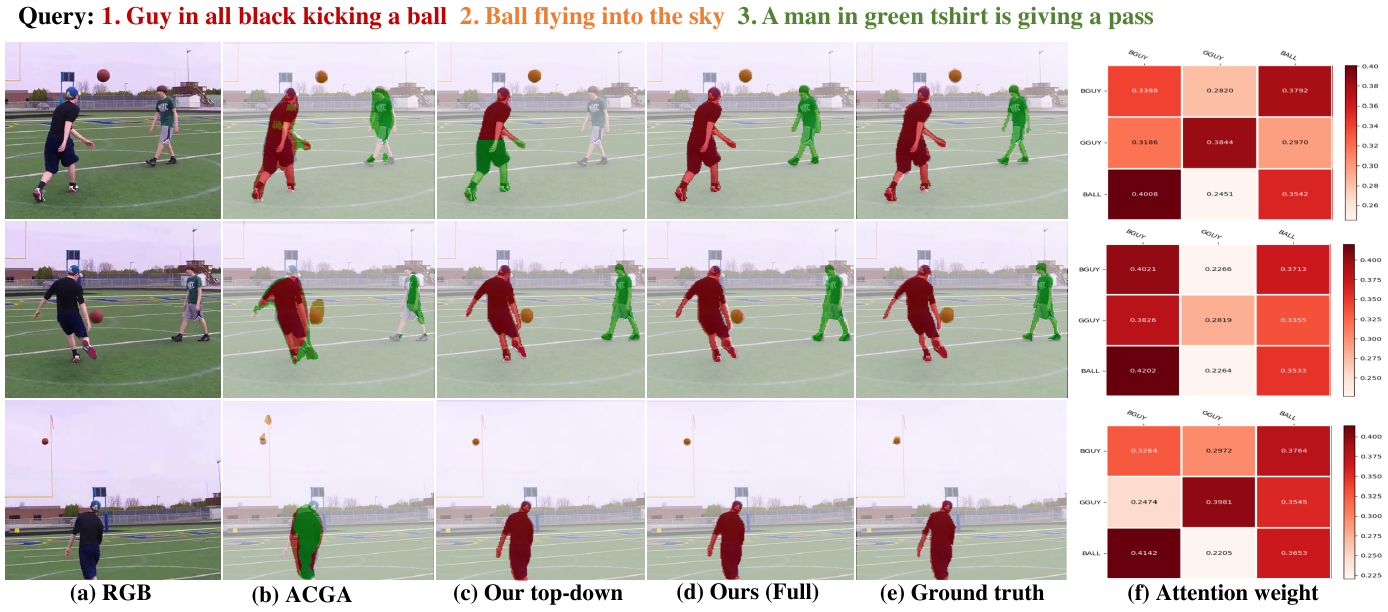}
    \put(-369,4){\scriptsize~\cite{wang2019asymmetric}}
\end{center}
\vspace{-16pt}
\captionsetup{font=small}
     \caption{\small\textbf{Qualitative results of text-based video segmentation.} We show three language query, and draw the corresponding segmentation results using the same query color.
     As queries 1 and 3 are predicted on a single object, the left most object in the first row of (c) is covered with both red (query 1) and green (query 3).
     (a) Original frames. 
     (b) Results of the bottom-up method~\cite{wang2019asymmetric}.
     (c) Results of our basic top-down pipeline (row 1 in Table \ref{table:ablation}). 
     (d) Results of our full model.
     (e) Ground truth.
     (f) Visualization of attention weights in our TSRM.
     }\label{fig:quan}
\vspace{-8pt}
\end{figure*}

\begin{figure*}[t]
\begin{center}
    \includegraphics[width=1\linewidth]{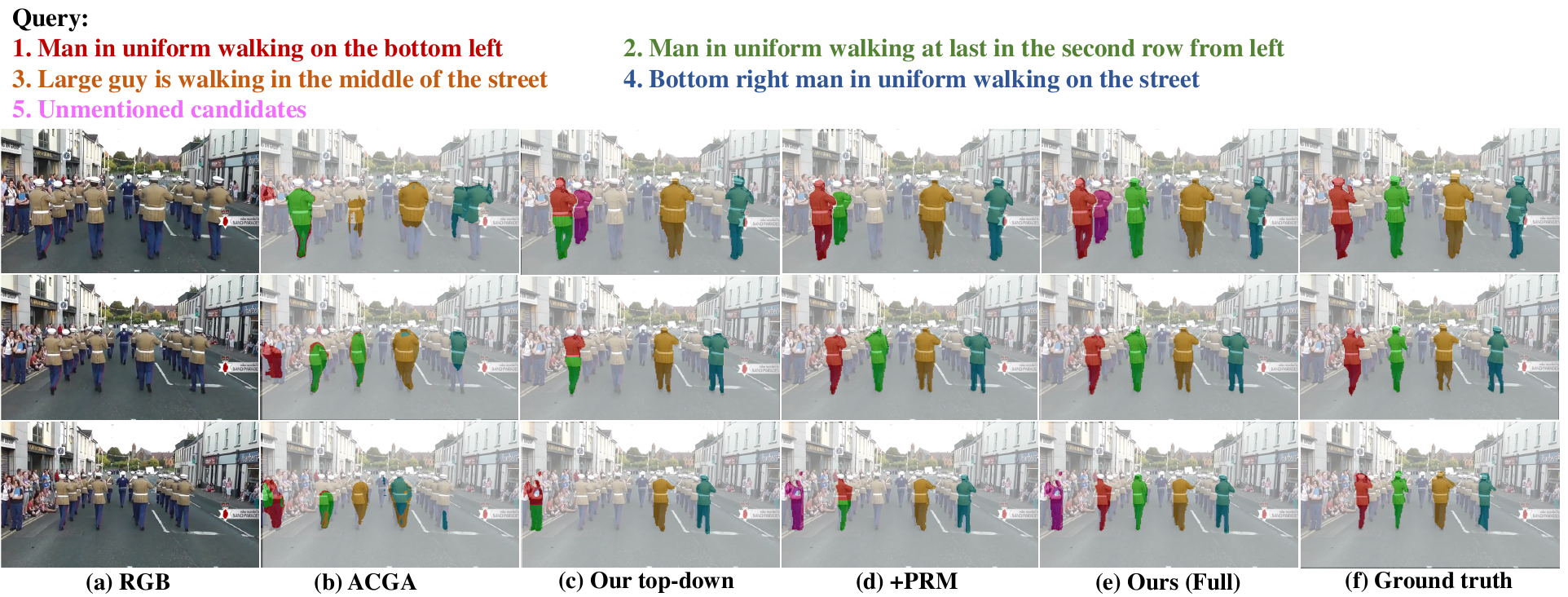}
    \put(-365,2){\scriptsize~\cite{wang2019asymmetric}}
\end{center}
\vspace{-16pt}
\captionsetup{font=small}
     \caption{\small\textbf{Visualization results of a complex video.} An object is covered with different colors if it is referred by more than one queries, \eg, the left most object in the first row of (c). 
     (a) Original video frame. 
     (b) Results of the bottom-up method~\cite{wang2019asymmetric}.
     (c) Results of our top-down pipeline.
     (d) Results of the PRM-enhanced top-down model (row 4 in Table \ref{table:ablation}).
     (e) Results of our full ClawCraneNet.
     }\label{fig:quan_2}
\vspace{-8pt}
\end{figure*}

\subsection{Ablation Studies}

\noindent\textbf{Effectiveness of Top-Down Pipeline.}
We first investigate the effectiveness of our designed Top-Down Pipeline.
We evaluated the basic top-down pipeline (segment-embed-retrieve pipeline), which ignores all the relation information among candidate objects and removes all relation-based modules.
The results are reported the results in Table~\ref{table:ablation}.
Compared to state-of-the-art \textit{bottom-up} methods shown in Table~\ref{table:a2dsota}, our \textit{top-down} pipeline achieves significantly better performances, especially on high precision predictions.

\noindent\textbf{Impact of Instance Segmentation Modules.}
In our experiments, for better trade-off between time cost and performance, we employ a one-stage segmentation method \cite{tian2020conditional}.
The performance of our ClawCraneNet with different instance segmentation methods is shown in Table \ref{table:ins}. 
We use CondInst(R-101-FPN) as the off-the-shelf instance segmentation module by default.
With weaker instance segmentation models, we still shows competitive performance compared with bottom-up approaches.

\noindent\textbf{Impact of Positional Relation Module.}
As shown in Table~\ref{table:ablation}, we have tried to adding different positional encoding methods, and achieved significant improvements compared to the basic top-down pipeline.
We can conclude that position information is very useful for the top-down framework of text-based video segmentation task.
In addition, we found our full Positional Relation Module (PRM) achieves better performances compared to absolute and relative position encoding methods. The reason is that absolute position encoding lacks the sensation for relative description like \textit{``the second from left"} and thus it is hard for exclusively relative encoding to balance the weights for formulating absolute information and relative information.

\noindent\textbf{Impact of Text-guided Semantic Relation Module.}
We further evaluate effectiveness of the proposed TSRM. As shown in Table~\ref{table:ablation} (5-6 row),
the vanilla object-level self-attention does not benefit the performance. 
But certain improvement occurs when introducing text guide to formulate object-level relations.
A possible reason is that the plain visual-based relation module cannot correctly gather relational information by just measuring context similarities. But with more linguistic information contained, concrete relations could be formulated, leading to a positive impact on the performance. 

\noindent\textbf{Impact of Temporal Relation Module.} By fully utilizing the temporal coherence, our ClawCraneNet with temporal relation module (the last row in Table~\ref{table:ablation}) further enhanced outperforms all the other models which validate the effects of the design.
Conclusively, these results confirm the merits of the object-level relation formulation again.

\subsection{Qualitative Analysis}

We would like to investigate the internal mechanism in ClawCraneNet by analyzing qualitative results.
Compared with the bottom-up method ((b) of Figure~\ref{fig:quan}, Figure~\ref{fig:quan_2}), our top-down pipeline shows reasonable segmentation results, while the other messes up the relational information and lead to ambiguous foreground masks. 
As shown in Figure~\ref{fig:quan} (c) and (d), when introducing the text-guided semantic relation module, our network learns to capture mutual information of corresponding objects. Visualization examples of text-guided attention weights are shown in Figure~\ref{fig:quan}~(f).
The comparisons between complete ClawCraneNet and alternative structures are illustrated in Figure~\ref{fig:quan_2}.
Only a part of the objects are labeled with language description in A2D Sentences, and 
we illustrate the unmentioned objects in Figure~\ref{fig:quan_2} with purple masks. 
Without the relative position module, the model tends to focus on objects that match the absolute position description ``left",  resulting in a wrong prediction. Ourtemporal relation module helps to achieve temporal consistency among frames, and correct the misunderstanding of the previous module (Figure~\ref{fig:quan_2} (e)).
In conclusion, these visualized results show the effectiveness of our top-down design and the object-relational modules in ClawCraneNet.

\section{Conclusion}
In this paper, we propose a novel ClawCraneNet following the segment-comprehend-retrieve strategy for the first attempt of introducing object-level relation into text-based video segmentation field.
Different from previous bottom-up methods,
our ClawCraneNet maximizes the semantic information flow between object-level features by fully investigating the relationship between intra-frame and inter-frame objects, \ie, positional relation, text-guided semantic relation, and inter-frame temporal relation. 
Evaluations on commonly used benchmark datasets demonstrate that ClawCraneNet surpasses all the state-of-the-art methods by large margins.

\clearpage

\appendix
\section{Appendix}
\subsection{Analysis of Puppet Mask in J-HMDB}
\begin{figure}[t]
\begin{center}
     \includegraphics[width=1\linewidth]{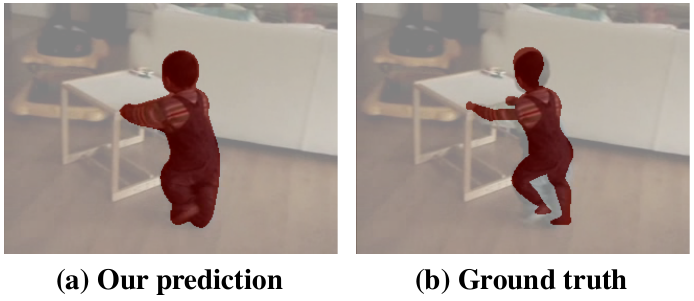}
\end{center}
\vspace{-8pt}
      \caption{
      Qualitative comparison between the predicted mask by ClawCraneNet and the ground-truth puppet mask.
      }\label{fig:puppet}
\end{figure}

In this section, we give a brief explanation about the poor performance of $P@0.9$ on J-HMDB Sentences \cite{jhuang2013towards} dataset (Line 8 in Table \textcolor[rgb]{1,0,0}{2}). 
As shown in Figure \ref{fig:puppet}, ground-truth masks from J-HMDB Sentences \cite{jhuang2013towards} are performed by puppets which leads to inconsistency between the segmentation mask and the actual object.
Since the evaluated model is trained from precise segmentation masks on A2D Sentences \cite{xu2015can}, it is hard for ClawCraneNet to fit the data distribution of J-HMDB without fine-tuning.

\subsection{Analysis of Object Embedding}
\begin{figure}[t]
\begin{center}
     \includegraphics[width=0.9\linewidth]{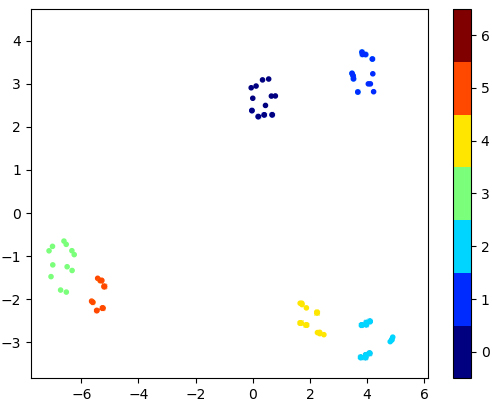}
\end{center}
\vspace{-8pt}
      \caption{
      T-SNE Visualizations of learned object embeddings from ClawCraneNet.
      }\label{fig:tsne}
\end{figure}

We visualize predicted embeddings of 12 language queries for all candidate objects on a randomly selected video in A2D Sentences \cite{xu2015can} validation set and use t-SNE \cite{maaten2008visualizing} to embed visual object embeddings (256-dim) into a 2D space. 
As shown in Figure \ref{fig:tsne}, embeddings belonging to different objects have clearly distinguishable margins, confirming that ClawCraneNet learns discriminative object embeddings which meets the prerequisites of temporal relation module.

\subsection{More Details about Predicted Results}
\begin{figure*}[t]
\begin{center}
    \includegraphics[width=.93\linewidth]{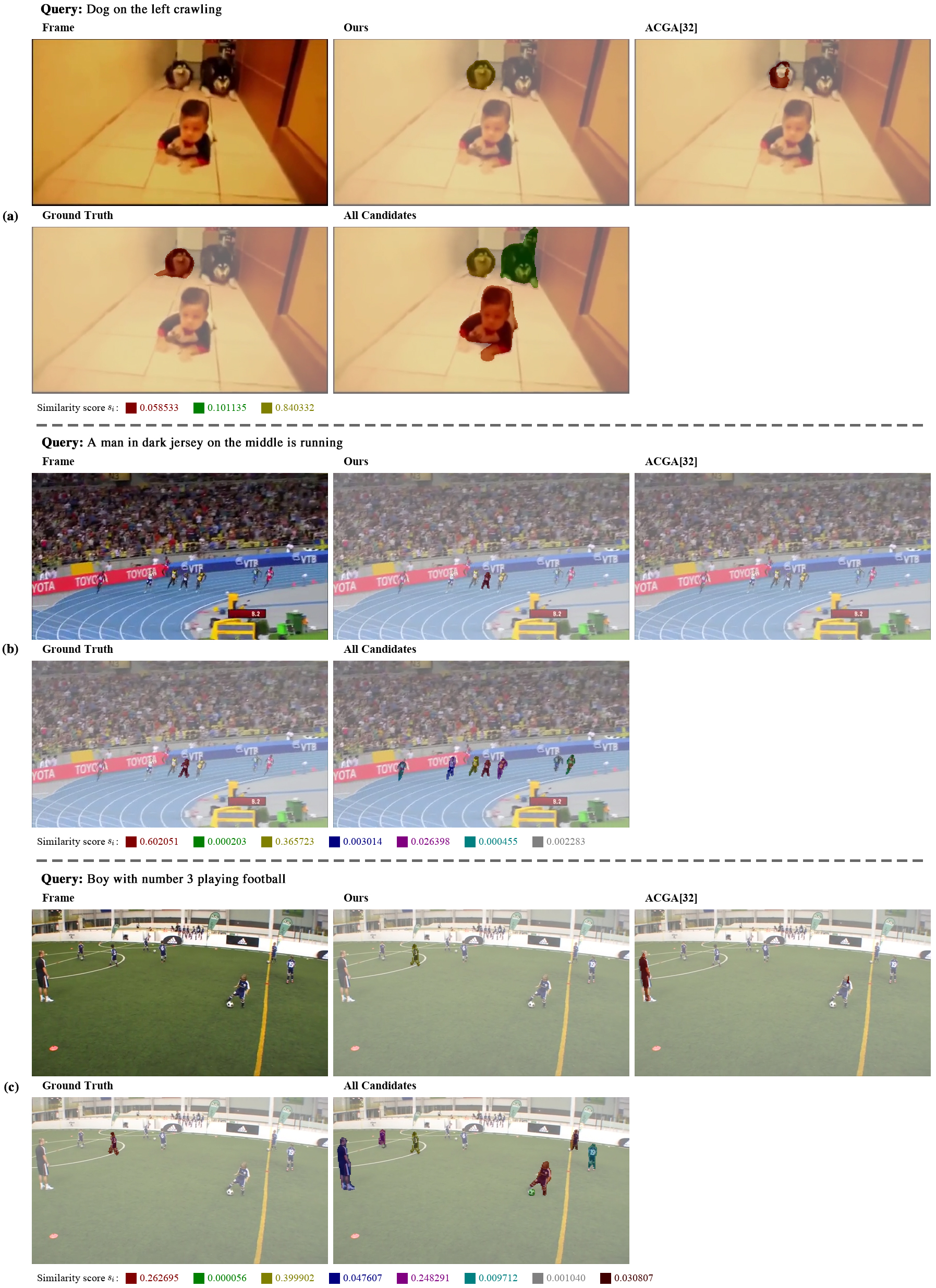}
\end{center}
\vspace{-0.15in}
     \caption{Qualitative results of text-based video segmentation.
     }\label{fig:perform}
\end{figure*}

In Figure \ref{fig:perform},  we detail the predictions of our ClawCraneNet. In each sub-figure, we give the original RGB frame and language query as input, then plot the predicted results of our ClawCraneNet and ACGA \cite{wang2019asymmetric}. \textit{All candidates} refers to candidate objects perceived by the instance segmentation module. 

\begin{figure*}[t]
\begin{center}
    \includegraphics[width=.93\linewidth]{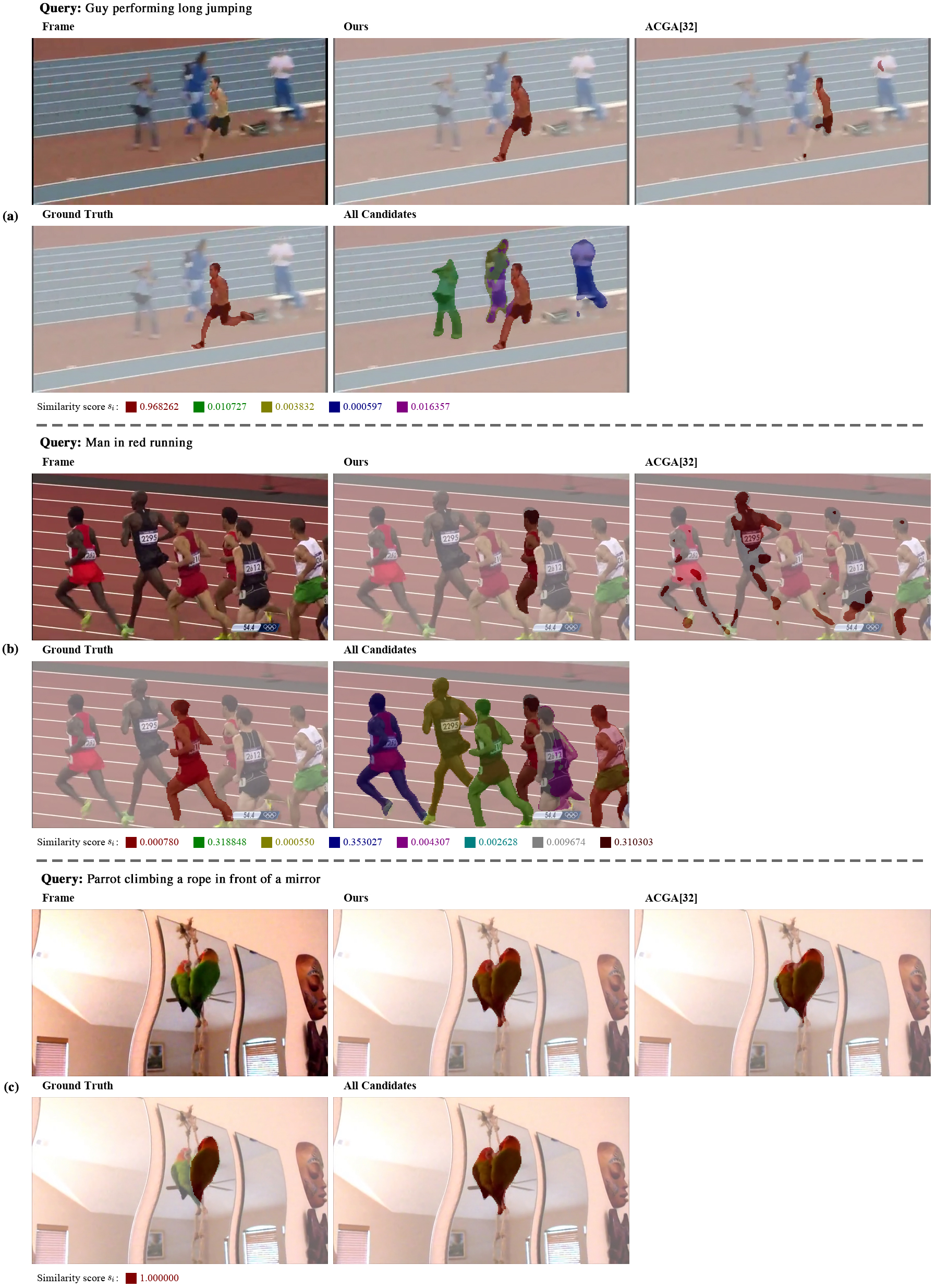}
\end{center}
\vspace{-0.15in}
     \caption{Failure cases of text-based video segmentation.
     }\label{fig:failure}
\end{figure*}

Thanks to the object-level comprehension of ClawCraneNet, reasonable results with clear boundaries are generated. Specifically, compared to bottom-up methods which mainly focus on salient objects, our network could entirely perceive inconspicuous objects and distinguish them with semantic context.
Some failure cases are shown in Figure \ref{fig:failure}, it is still hard for ClawCraneNet to handle some visual ambiguity, \ie, objects moving at high speed (Figure \ref{fig:failure} (a)) or objects in the mirror (Figure \ref{fig:failure} (c)). 

An interesting observation is that with ambiguous language description like \textit{Man in red running}, multiple fitting objects can be highlighted with higher similarity scores, as shown in Figure \ref{fig:failure} (b).
To give more examples, we supply some results in form of videos within the supplementary material.

\clearpage

{\small
\bibliographystyle{ieee_fullname}
\bibliography{egbib}
}

\end{document}